\documentclass[12pt]{article}

\usepackage{amssymb}

\usepackage{hyperref} 

\usepackage[autostyle=true,german=quotes]{csquotes} 

\usepackage{amsmath,amssymb,amsthm,amsfonts,amsbsy,latexsym}

\usepackage{caption}
\usepackage{subcaption}

\usepackage{algorithmic}
\usepackage[algoruled,german]{algorithm2e}

\usepackage{graphicx}

\usepackage{tikz}
\usepackage{tabulary}

\usepackage{multirow}
\usepackage{array}
\usepackage{hhline}
\usepackage{xcolor}
\usepackage{booktabs}
\usepackage{longtable}
\usepackage{siunitx} 
\usepackage{csvsimple}

\usepackage[normalem]{ulem}
\usepackage{listings}

\definecolor{codegreen}{rgb}{0,0.6,0}
\definecolor{codegray}{rgb}{0.5,0.5,0.5}
\definecolor{codepurple}{rgb}{0.58,0,0.82}
\definecolor{backcolour}{rgb}{0.95,0.95,0.92}

\lstdefinestyle{mystyle}{,   
    commentstyle=\color{codegreen},
    keywordstyle=\color{codegreen},
    numberstyle=\tiny\color{codegray},
    stringstyle=\color{codepurple},
    basicstyle=\ttfamily\scriptsize,
    breakatwhitespace=false,         
    breaklines=true,                 
    captionpos=b,                    
    keepspaces=true,                 
    numbers=left,                    
    numbersep=4pt,                  
    showspaces=false,                
    showstringspaces=false,
    showtabs=false,                  
    tabsize=2
}

\usepackage[nameinlink,capitalise,german]{cleveref}

\newcommand{\R}{\mathbb{R}}

\usepackage{mathtools}

\DeclareMathOperator*{\argmin}{arg\,min}

\usepackage{color}

\usepackage[nottoc]{tocbibind}

\usepackage{natbib}
\usepackage{hyperref}                   
\hypersetup{                            
    colorlinks=true,                    
    linkcolor=black,                     
    citecolor=black,                     
    urlcolor=black                       
}

\title{Churn modeling of life insurance policies via statistical and machine learning methods - Analysis of important features}
\author{Andreas Groll\thanks{Department of Statistics, TU Dortmund University; corresponding author: groll@statistik.tu-dortmund.de}, Carsten Wasserfuhr\thanks{Department of Statistics, TU Dortmund University}, Leonid Zeldin\thanks{Department of Statistics, TU Dortmund University}}
\date {February 2022}

\begin{document}

\maketitle

\mbox{}
\vfill

\noindent Statements relating to the ethics and integrity policies: \\~\\

\noindent Unfortunately, our data are confidential and not available. However, if requested by the referees, we could make example code available, which illustrates the usage of our fitting procedures. Furthermore, we have no funding and no conflict of interest to declare.
\pagebreak

\begin{abstract}


\noindent 
Life assurance companies typically possess a wealth of data covering multiple systems and databases. These data are often used for analyzing the past and for describing the present. Taking account of the past, the future is mostly forecasted by traditional statistical methods. So far, only a few attempts were undertaken to perform estimations by means of machine learning approaches. In this work, the individual contract cancellation behavior of customers within two partial stocks is modeled by the aid of various classification methods. Partial stocks of private pension and endowment policy are considered. We describe the data used for the modeling, their structured and in which way they are cleansed. The utilized models are calibrated on the basis of an extensive tuning process, then graphically evaluated regarding their goodness-of-fit and with the help of a variable relevance concept, we investigate which features notably affect the individual contract cancellation behavior. 


\end{abstract}


\section{Introduction}

%
%

\noindent Data are among the biggest treasures of the 21st century. Financial service providers like insurers 
hold large data quantities which could provide a significant improvement of the comprehension of their 
business if exploited correctly. Life insurance as one of the most important and oldest sectors of insurance is 
characterized by its long-term contracts. Inventory management turns out to be a very central aspect for better 
understanding and controlling the existing contracts in the company.
An important goal is the creation of short- and long-term strategies, including the evaluation of solvency, 
as it is frequently examined by various surveillance authorities. One sub-aspect of the inventory management 
is the contract reversal. Being able to comprehend this aspect and its causes can lead to various actions like the reversal prevention.

Due to the rapid development of inventory management systems and the increase of computational power, 
many applications of {\em Big Data} became more and more relevant in recent years.
Along with this, powerful and advanced methods of statistical and machine learning have been developed, 
which allow to analyze large data quantities, to recognize connections within the data and to predict future 
events.  Lately, machine learning techniques have become more and more important 
in the context of modeling insurance data, see e.g.\ \citet{MAKARIOU2021140, DEVRIENDT2021248, DENUIT2021485, GAN2013795}. In this work, some of these methods shall be used for a better understanding of the individual 
contract cancellation behavior of a large German life insurer's clients. For this purpose, classical statistical 
approaches like the logit regression model \citep{fahrmeir2001models, mccullagh2019generalized} and 
the elastic net regularization method \citep{elastic_net} as well as modern machine learning methods are 
employed. From the field of machine learning, several tree-based approaches are considered. These 
include Classification and Regression Trees (CART; \citealp{breiman_cart}), Random Forest  
\citep{breiman_random} and XGBoost \citep{chen2016xgboost}. 

In this manuscript, first we will briefly introduce the previous methods for statistical modeling 
as well as various evaluation approaches for the models. These are then applied to the partial 
stocks of endowment life insurance (ELI) and of private pension insurance. For each approach 
the respective best model regarding the so-called {\it Area Under the Curve} (AUC; see, e.g., 
\citealp{bradley1997use}) is determined by means of parameter tuning. Subsequently, the resulting 
models are investigated with respect to the relevant factors for decision making. These factors shall 
give information on the relationship between the contract reversal and various contract parameters 
and individual client information. 

In Section~\ref{sec:motiv}, we briefly describe the underlying research problem and motivate why it 
is useful to understand individual contract termination behavior in more detail.
Section~\ref{sec:data} includes a description of the database used for model building. We describe 
the general structure of the data as well as key aspects of data preprocessing.
Then, in Section~\ref{sec:methods}, we briefly describe the different approaches used for model building, 
their main aspects, and how these models are evaluated. Specifically, in Section~\ref{subsec:var_rel_theor}, 
we describe in detail a certain measure of {\it Varibale Relevance} defined in this work.
Next, we present the main results in Section~\ref{sec:results}, comparing the different approaches and their 
corresponding tuned models using the methods described in Section ~\ref{sec:methods}. 
Finally, we summarize the results in Section~\ref{sec:conclusion} and give an outlook on how this work can be extended.

\section{Motivation}\label{sec:motiv}

\noindent Life insurance differs from other types of insurance in that it offers products with a very long term. 
Particularly, in term risk life insurance, endowment life insurance and occupational disability insurance, high 
premiums are paid in the beginning for acquisition costs and administrative expenses. Over time, the initial costs 
incurred are compensated by the premiums and the associated investment income. Subsequent to this period, 
profits are generated. 

In order to create a stable financial basis in the long term, it is necessary to address individual contract cancellation behavior. 
Although this subject is well researched in the deterministic field of actuarial science, to the best of our knowledge,
there are few analyses that use statistical techniques to predict future lapse-related charges. Indeed, it is of great advantage 
for an insurance company to be able to anticipate certain lapse rates, because then suitable measures could 
be taken to prevent lapses. 

This manuscript deals with a number of statistical and machine learning estimation methods which attempt to investigate 
the lapse behavior of individual contracts of the portfolio of a large German life insurer on a large data basis, and to compare 
them with respect to their predictive validity. All calculation have been performed in the statistical software program \citet{R}.

\section{Data}\label{sec:data}
\noindent In order to build up the estimation methods for explanation and prediction of the cancellation behavior of individual contracts, a sufficiently large and high-quality data base is required. In the context of this work, we were able to access a large amount of information from the data warehouse (DWH). 
It contains an extensive collection of information on insurance contracts and insured persons, which in turn is grouped according to various logics. 

Initial considerations about which information is relevant for investigating lapse behavior in individual contracts 
led to the conclusion that the most suitable data basis is portfolio data on the basis of contract information. 
This means that, in addition to information on the main insurance policy, information on all associated supplementary 
insurance policies is also available. The period 1/1/2018-12/31/2018 was selected as the study period. 

Each insurance contract is assigned an anonymous contract number within the DWH.
The contract part ID can then be used to assign the corresponding contract part. 
Hence, a contract part of an insurance can be uniquely assigned by the combination of both 
characteristics within the DWH. An example of this structure is displayed in Table~\ref{fig:aufbau_daten_dwh}.
Since these two pieces of information are unique, but also randomly filled, they are only important 
during data preparation and are neglected for model building. 
 
\begin{table}[h]
\begin{tabulary}{8cm}[h]{ccccc}
  contract ID & contract part ID & $\ldots$ & sum insured & actuarial interest rate  \\
 \midrule
  1678655 & 1 & $\cdots$ & 48.250 & 3,25 \\
 1678655 & 2 & $\cdots$ & 37.630 & 2,75  \\
 4789889 & 1 & $\cdots$ & 18.470 & 1,25  \\
 4912002 & 1 & $\cdots$ & 16.800 & 1,25  \\
 5100200 & 1 & $\cdots$ & 4.520 & 0,9  \\
 $\vdots$ &$\vdots$ &$\vdots$ &$\vdots$ &$\vdots$  \\ \midrule
\end{tabulary}	     
\caption{Structure of the inventory data. In this example, four contracts are shown, as indicated by the contract number. 
The first contract contains two contract parts, the others only the main insurance. The contract parts contain, among other things, 
the information on the sum insured and the actuarial interest rate.}  
\label{fig:aufbau_daten_dwh}
\end{table}

Moreover, for an analysis of the cancellation behavior of individual customers, statistical information on the
annual premium, the sum insured, and the actuarial interest rate associated with the contract, which among 
other things enables the contract to be classified on the time axis, can also be expected to be of great importance. 

Another important data source is collection data. The prior events listed there can be classified into eight categories 
on the basis of one feature that describes the type of collection process, such as, e.g., delay of premium payments 
and/or resulting dunning procedures. For each insurance contract and each of these eight categories, the number of prior events 
within the last five years (for endowment insurance) or within the last year (for private pension insurance) is then calculated.
The distinction between these two types of insurance could be made on the basis of the name of the contract part.   

Besides the acquisition of the data, their preprocessing was an essential part of the preparatory work, 
which had to be carried out in advance of our analyses. The preprocessing is divided into some central 
steps, which are described in the following.

\subsection{Feature Engineering} \label{subsec: feature engineering}

\begin{figure}[h]
\includegraphics[width=\linewidth]{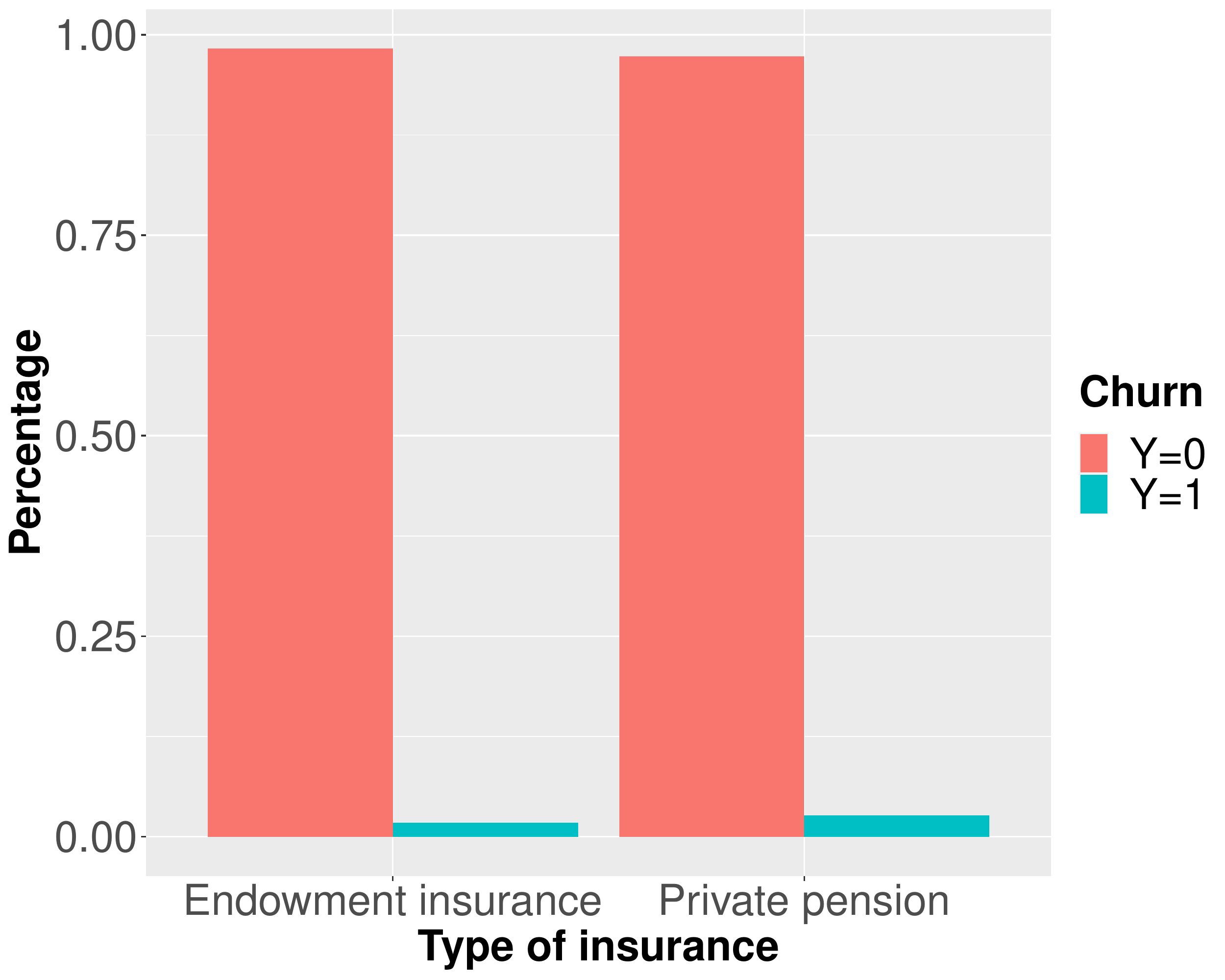}
\caption{Proportions of canceled and non-cancelled contracts within the data sets under review.}
\label{fig:storno_anteile}
\end{figure}

\noindent One of the most important features that had to be created is the binary target variable 
\enquote{individual contract cancellation}. Here, the reason for the contract's termination was considered: 
If it was a canceled contract, the variable was assigned the value 1; if the contract had another or no 
reason for termination, the variable was assigned the value 0. Since it is common for life insurance 
products to have a rather long duration, only very few contracts in the respective final data set have the value 1 
in the target variable ``individual contract cancellation", see Figure~\ref{fig:storno_anteile} and Table~\ref{tab:FE_daten_dwh}. 
Hence, the data set is \enquote{unbalanced} with respect to the target variable. We shall describe in the following how this can be handled. 

\begin{table}[h]
\begin{tabulary}{13.7cm}[h]{RCCCCCCC}
 & contract ID & contract part ID & $\ldots$ & contract beginn & occupation & $\ldots$ & contract cancelled? \\
\midrule
1: & 1678655 & 1 & $\cdots$ & 01.02.2001 & Baker & $\cdots$ & 1 \\
2: & 1678655 & 2 & $\cdots$ & 01.04.2005 & Baker & $\cdots$ & 0 \\
3: & 4789889 & 1 & $\cdots$ & 01.07.2015 & Police Officer & $\cdots$ & 0 \\
4: & 4912002 & 1 & $\cdots$ & 01.09.2015 & Craftsman & $\cdots$ & 0 \\
5: & 5100200 & 1 & $\cdots$ & 01.03.2017 & Actuary & $\cdots$ & 0 \\
$\vdots$ & $\vdots$ &$\vdots$ &$\vdots$ &$\vdots$ &$\vdots$ &$\vdots$ &$\vdots$ \\
\midrule
\end{tabulary}	     
\caption{Extension of the inventory data with self-built features. The contracts from 
Figure~\ref{fig:aufbau_daten_dwh} are extended by, among others, the contract start date, 
the categorized occupation and the target variable individual contract cancellation.}
\label{tab:FE_daten_dwh}
\end{table}      

From the temporal information regarding the beginning and expiration of the insurance, the information of 
the total, already passed and remaining contract duration was calculated and added to the data set. 
Furthermore, the current age, as of 1/1/2018, of the first insured person was calculated by means of the 
so-called {\it semi-annual method}\footnote{An actuarial calculation method; for this purpose, the exact age of the 
insured person at the start of the contract is rounded off commercially and extrapolated accordingly, 
see \citet{halbjahresmethode}. In actuarial terms, one is six months before and after the birthday as old as on the birthday itself}.  

Based on the information on the existence of additional insurance policies further (dummy) features were created, 
indicating whether a specific product type was chosen as an additional insurance policy. In addition, 
the information on the number of rejected dynamic policies\footnote{A dynamic policy, or more 
precisely a dynamic premium policy, which allows the customer to dynamically increase premium payments 
according to a certain time plan, see \citet{halbjahresmethode}} was added to the database.

The categorical covariate from DWH, which contains the information on the occupation of the first insured person,
has far too many different levels, such that the implemented learning procedures cannot handle this.
Therefore, a coarser categorization of the occupation status was developed, which was then incorporated into the 
data set of the private pension. 
            
Similar to the case of the occupation status, the information about the sales of the insurances 
had too many categories. For this reason, the general designation of the national directorate was used, 
which turned out to be well suited for the construction of models.          

\subsection{Selections}

\noindent First, several features were removed from both data sets in advance, as they were either not filled 
or filled very sparingly or were not relevant in terms of content. Furthermore, endowment life insurance (ELI) 
contracts with a death date or an expiration date prior to January 1st, 2018, were removed. 
On the one hand, it was important for the private pension insurances that they may not be in 
the status of benefit and, on the other hand, that they are pure private pension insurances. Only contracts that 
meet these requirements were used for the following analyses.

An important selection made after processing the raw data is the distinction between main and supplementary 
insurance, since we only examine the cancellation behavior of main insurance policies. Since further information is generated 
beforehand on the basis of possible additional insurances, this selection step must be carried out at the end of the preprocessing. 

\subsection{Imputations}
\noindent For contracts where the customer does not pay regular premiums, but has only paid 
a single premium, there is no information about the annual premium. 
Therefore, these missing values have been replaced by the value 0. 
            
In the case of ELI, some information was missing regarding gender and date of birth of the 
first insured person. Hence, an attempt was made to fill in the missing information by 
comparing the information on the policyholder, since the person who pays for the contract usually 
also claims the insurance coverage. In most cases, this approach was successful. 
            
The information generated in Section~\ref{subsec: feature engineering} about the different durations 
as well as the start and end time of the contract partly exhibit a strong positive correlation. 
For this reason, some features like the information about already passed contract duration were finally removed again.

\section{Methods}\label{sec:methods}
\noindent In the following, we give an overview of the different approaches used to model churn of life 
insurance policies and explain how these approaches are evaluated. In particular, the concept of 
{\it variable relevance} is discussed in detail.
\subsection{Classification Models}
\noindent To investigate the research question mentioned in the introduction, five different types 
of models were used, firstly logistic regression approaches and secondly tree-based methods.\\
\subsubsection{Logit Model}
\noindent The goal of a regression model with a binary target variable is to model and estimate the effects of the variables on the probability
            \begin{align*}
                \pi &= P(y = 1) = P(y = 1 \mid x_1,\ldots,x_p) \\
                    &= E(y \mid x_1,\ldots,x_p) ,
            \end{align*}
            i.e. the occurrence of the event $y=1$ given the realizations of the variables $x_1,\ldots,x_p$.
            This is implemented with the help of a Generalized Linear Model (GLM).
            For this purpose, we first define a linear predictor
            \begin{align*}
                \eta = \beta_0 + \beta_1 x_1 + \ldots + \beta_p x_p = \pmb{x}^{\top} \pmb{\beta}
            \end{align*}
            with $\pmb{\beta} = \left( \beta_0, \beta_1, \ldots , \beta_p \right)^{\top}$ and $\pmb{x} = \left( 1, x_1, \ldots , x_p \right)^{\top}$. 
            Using the logistic response function 
            \begin{align}
                h(\eta) = \frac{\exp{(\eta)}}{1+\exp{(\eta)}}, \label{eq:logit_response}
            \end{align}
            the probability $\pi$ is then calculated as follows:
            \begin{align*}
                \pi = \pi_{\pmb{\beta}} = h(\eta) = h(\pmb{x}^{\top} \pmb{\beta}).
            \end{align*}
A model that uses the response function \eqref{eq:logit_response} is called a {\it logit model}. 
Similar to the ordinary linear model, in this model the $\beta$ coefficients represent the covariate effects and 
can be interpreted quite straightfowardly.

The $\beta$ coefficients are computed using the Bernoulli loss by minimizing the log-likelihood. 
Further details on GLMs in general can be found in \citet{mccullagh2019generalized}.

\paragraph*{Elastic Net}
\noindent Similar to the calculation of the $\beta$ coefficients in an ordinary linear model, 
the estimates of the coefficient vector $\pmb{\beta}$ in the logit model are often unstable and lead to high variances. 
This occurs, for example, if there are substantial correlations between covaroates \citep{regression_fahrmeir_et_al}.
In such cases often regularization techniques are applied to obtain more appropriate values for the 
parameters $\pmb{\beta}$, e.g.\ the LASSO, see \citet{tibshirani1996regression} and \citet{elastic_net1}.
                
In this work, we use the elastic-net approach of \citet{elastic_net}, which is a combination of 
the LASSO and ridge regression \citep{hoerl1970ridge}. Let 
                \begin{align*}
		            pen_\alpha(\pmb{\beta}) = \alpha \| \pmb{\beta} \|_1 + (1-\alpha) \frac{\| \pmb{\beta} \|_2}{2}
		        \end{align*}
be a penalty term with $\alpha \in [0,1]$, which measures the complexity of the parameter vector $\pmb{\beta}$, and $\lambda > 0$ a 
regularization parameter, which controls the trade-off between the fidelity of the estimation of $\pmb{\beta}$ and the influence of the penalty.
An estimate of $\pmb{\beta}$ is obtained by solving the optimization problem
                \begin{align*}
		            \hat{\pmb{\beta}} = \underset{\pmb{\beta} \in \R^{(p+1)}}{\argmin} ~\lbrace \ell(\pmb{\beta}) + \lambda \cdot pen_{\alpha}(\pmb{\beta}) \rbrace, 
		        \end{align*}
where $\ell(\pmb{\beta})$ is the log-likelihood of the logit model.
For $\alpha=0$, this is equal to ridge regularization, while for $\alpha=1$ the LASSO is obtained.
		        
\subsubsection{Tree-based Models}  
        
\noindent Tree-based models are a class of popular models that are often very powerful in terms of prediction.
The basis of tree-based classification methods is the so-called classification tree, which is also easy to interpret.
In this work, we use an implementation of the classification trees based on the {\it classification and regression trees} 
(CARTs) of \citet{breiman_cart}.

\paragraph*{CARTs}
\noindent The basic idea of CARTs is the recursive partitioning of the input space, and then computing a 
prediction per partition of the input space according to certain rules. The main goal of this method is to find 
the best possible partitions, so that the observations in a partition are as similar as possible with respect to the target variable.
	            
Classification trees have a simple structure, are easy to visualize, and, hence, can be easily interpreted.
Unfortunately, CART models often exhibit a high variance, which is why extensions 
of the CART model, so-called ensemble models, are often used.
These ensemble methods are usually divided into {\it bagging} and {\it boosting methods}.
One bagging approach used in this work is the so-called {\it random forest} of \citet{breiman_random}. 
Furthermore, the so-called {\it extreme gradient boosting} (XGBoost) method will also used.

\paragraph*{Random Forest}
\noindent The basic idea of the random forest is to combine several classification trees.
For this purpose, single classification trees are computed independently from each other 
and then the official predictions of all computed trees are combined into a single prediction by a majority vote.
                
 In order to reduce the variance of a random forest significantly compared to a single tree, 
 the trees should be as uncorrelated as possible. This is achieved by adding randomness to the 
 construction of the single classification trees. Thus, firstly the single trees are drawn on (random) bootstrap samples 
 from the original data set. Furthermore, only a random selection of variables is allowed to compute the 
 best split at each node, see \citep{breiman_random}. Pruning of the computed tree is also omitted. The 
 combination of de-correlated trees results in predictions that achieve smaller bias and variance than 
 those of a single CART model.

\paragraph*{XGBoost}
\noindent    The {\it eXtreme Gradient Boosting} (XGBoost) by \citet{chen2016xgboost} is an 
extension of the {\it gradient tree boosting} method by \citet{friedman2001greedy}, which achieves 
very good prediction results in practice, see \citet{xgboost_award}, and is currently one of the 
\enquote{state-of-the-art} learning methods from the machine learning field \citep{bentejac2021comparative}.
The basic principle of boosting in general is to construct a \enquote{strong} 
learning procedure by sequentially combining many \enquote{weak} learning procedures.
In the following, classification trees are used as base learners.
                
Besides a sophisticated implementation, the XGBoost method uses a regularized risk function to 
compute a classification tree. This risk function is a regularized version of the empirical risk and is defined as
\begin{align*}
\mathcal{R}_{reg}^{[m]} = \sum_{i=1}^n L \left( y_i, f^{[m-1]}(x_i) + f^{[m]}(x_i) \right) + J\left( f^{[m]} \right)
\end{align*}
with
\begin{align*}
J\left( f^{[m]}\right) = \lambda \cdot J_1 \left( f^{[m]}\right) + \gamma \cdot J_2 \left( f^{[m]}\right) + \alpha \cdot J_3 \left( f^{[m]}\right).
\end{align*}		            
Here, $J\left( f^{[m]}) \right)$ describes a penalty term of the $m$-th constructed tree and 
controls the tree structure of the computed tree models. $J_1$ controls the tree depth based 
on the number of terminal nodes. $J_2$ and $J_3$ control the prediction to be computed for the 
final knots of the $m$-th constructed tree using $L_2$ and $L_1$ regularization, respectively.
		            
In addition to the use of a different risk function, the computation of the $m$-th constructed tree is 
approximated around the value $f^{[m-1]}(x)$ using a second-order Taylor series. 
On the one hand, this allows for simplified calculations in the context of risk minimization, and on the other 
hand parallel calculations of the node splits. Further details can be found in \citet{chen2016xgboost}.

\subsection{Unbalanced Target} \label{sec:unbalanced_data}
\noindent One of the central problems of the analyzed data sets is the strong imbalance of the binary target variable. 
As shown in Figure~\ref{fig:storno_anteile}, the target variable is strongly unbalanced in both data sets.  

Data sets with an unbalanced target variable have the property that corresponding prediction models are biased in favor of the majority class.
This means that observations of the minority class are often predicted incorrectly, while the overall misclassification rate is still low.

Thus, evaluating a model trained on highly unbalanced data by the {\it misclassification rate} is often inappropriate.
Therefore, for unbalanced classification problems, among other things, more specific goodness-of-fit criteria than 
the misclassification rate are needed, which will be discussed in more detail in Section~\ref{subsec:evaluation}.
		
However, note that the use of other goodness-of-fit measures only changes the description and interpretation of the results 
of a model, but not the training process itself. Therefore, in addition to the use of more appropriate goodness-of-fit measures, 
an oversampling method was used to change the structure of unbalanced data sets during the training of the models.
The goal of this method is to increase the size of the minority class, resulting in a training data set that is more balanced 
with respect to the target variable. This is achieved by randomly adding copies of single positive observations 
(here, observations with $y_i=1$) and is called {\it random oversampling}.
		
A widely used alternative to this is the {\it Synthetic Minority Oversampling Technique} (SMOTE) by \citet{SMOTE}.
This method has also been tested by the authors throughout their analyses, but has not been shown to be better in this context.
		
\subsection{Evaluation} \label{subsec:evaluation}
\noindent Many methods for determining the goodness-of-fit of a classification model with a binary target 
variable are based on metrics using a confusion matrix. A confusion matrix compares the true and predicted 
classes calculated by a classification model in a contingency table, see Figure~\ref{fig:Konfusionsmatrix}.
For this purpose, a classification model determines a score for a given observation, based on which 
the observation is assigned to a concrete class with the help of a threshold. Specifically, such a binary assignment 
is defined such that if the score of an observation exceeds the threshold, it is assigned to a class one, and to class zero otherwise. 
Usually, a threshold of $0.5$ is used. The confusion matrix shows how many observations are correctly predicted 
by the given model (True Positive (TP) and True Negative (TN)) or falsely predicted (False Positive (FP) and False Negative (FN)).
As explained in Section~\ref{sec:unbalanced_data}, for problems with unbalanced data sets 
other goodness-of-fit criteria are required to evaluate models.

\begin{figure}[tp]
\centering
\bgroup
\def\arraystretch{2}
\hskip-1.0cm
\begin{tabular}{>{\centering}p{1cm}>{\centering\arraybackslash}p{2cm}|>{\centering\arraybackslash}p{2cm}|>{\centering\arraybackslash}p{2cm}||>{\centering\arraybackslash}p{2cm}|}
\multicolumn{2}{c}{}    &   \multicolumn{2}{c}{Prediction} & \multicolumn{1}{c}{}\\
&            &   Negative &   Positive   & Sum      \\
\cline{2-5}
\multirow{2}{*}{\rotatebox[origin=c]{90}{Observatino}} & Negative   &  $TN$   &   $FP$   & $N$             \\
\cline{2-5}
 & Positive  &  $FN$     & $TP$   & $P$             \\ 
\hhline{~====}
 & Sum   & $\hat{N}$     &  $\hat{P}$   & $n$             \\ 
\cline{2-5}
\end{tabular}
\egroup
\caption{Schematic representation of a general confusion matrix in the case of a binary target variable. 
The structure is based on the representation of the confusion matrix in the R-package \texttt{mlr} \citep{mlr}}
\label{fig:Konfusionsmatrix}
\end{figure}

In order to describe more suitable goodness-of-fit criteria, we first define further key quantities using the confusion matrix:
\begin{align*}
\text{Recall}  &=  \frac{TP}{P}, & \text{FPR}  &=  \frac{FP}{N}, \\
\text{Precision}  &=  \frac{TP}{\hat{P}}, & \text{TNR}  &=  \frac{TN}{N}.
\end{align*}
The quantities Recall\footnote{also called $TPR$, true positive rate}, FPR\footnote{false positive rate} and 
TNR\footnote{true negative rate} indicate for each class of the target variable how well or poorly the own class is predicted. 
The quantity Precision\footnote{also called $PPV$, positive predictive value} provides 
information on how many of the observations predicted as positive are actually positive. 		

\subsubsection*{F1-Measure}
\noindent  The so-called F1-measure is defined as the harmonic mean of Precision and Recall:
\begin{align*}
F1 := 2 \cdot \frac{Precision \cdot Recall}{Precision + Recall}\,.
\end{align*}
The focus of this goodness-of-fit measure is on the analysis of the minority 
class \citep{he2013imbalanced} by analyzing the relationship between the correctness 
and completeness of the prediction \citep{LearningfromImbData}.
It is $F1\in[0,1]$, where a value of 1 is the best possible value.

\subsubsection{ROC-Measures}
\noindent In the following paragraph, we explain the mainly used evaluation techniques, the 
{\it Receiver Operating Characteristics} (ROC) measures. ROC-measures describe ways to represent 
results of binary classification problems \citep{roc_pr_relationship}, compare models 
with each other, and graphically visualize their performance \citep{roc_intro}. 
The basis of the ROC-measures is the fact that classification models provide a probability of class membership 
for each observation and class. If the probability exceeds a certain threshold, the observation is predicted to be positive. 
Consequently, different predictions are dervied for different thresholds, resulting in different confusion matrices. 
ROC-measures take up this basic idea that the performance of a prediction or a model changes for different thresholds.

\subsubsection*{ROC-Curve}
\noindent For each threshold value, the Recall and the FPR can be calculated.
If the corresponding value pairs $(FPR,Recall)$ are plotted in a coordinate system, this results in a set of points, 
each representing a specific threshold value. Connecting these points as a step function 
results in a curve, the so-called {\it ROC curve}.
		        
Such a ROC curve represents the performance of one specific classification model.
A ROC curve can be interpreted in the sense that the more the ROC curve tends 
towards the point $(0,1)$, the better the model.

\subsubsection*{Precision-Recall-Curve}
\noindent  The Precision-Recall-curve, or PR curve for short, is a useful alternative or 
extention to the ROC curve, as it can show performance differences that remain undetected in an ROC curve \citep{improve_pr_curves}.
The basic idea of a PR curve is similar to that of a ROC curve, i.e.\ one wants to graphically display the model 
performance for different threshold values. In contrast to the ROC curve, the values $Recall$ and 
$Precision$ are considered here. Connecting the resulting points yields the PR curve.
The closer the PR curve is to the point $(1,1)$, the better the model.  
                
An important difference to the ROC curve is that the focus of a PR curve is on the correct prediction of positive observations. 
Deliberately, the number of correctly predicted negative observations $(TN)$ is not considered.
                               
 For unbalanced classification problems, PR curves are better suited than ROC curves, since ROC 
 curves in these cases often take an overly optimistic view of model performance, and PR curves also 
 provide more information about model performance with respect to the (underrepresented) 
 positive class, see, e.g., \citet{he2009learningfromimbdata} and \citet{roc_pr_relationship}. 

\subsubsection*{AUC}
\noindent   While the ROC curve is mainly a graphical representation of the goodness-of-fit 
of a model, an accurate comparison of multiple models requires a suitable metric \citep{LearningfromImbData}, 
since a comparison using ROC curves is not always clear due to its two-dimensionality \citep{roc_intro}.
A frequently used method in this context is the {\it Area Under the ROC Curve}  (AUC), 
see \citet{bradley1997use} and \citet{hanley1983method}.
The value of the AUC is calculated as the integral under the associated ROC curve 
and is practically calculated using the ``trapezoidal rule''.
The AUC can principally take values between 0 and 1, where 1 is the best possible AUC value.
Usually, however, the AUC takes values between 0.5 (random prediction) and 1 (perfect prediction).

\subsection{Variable Relevance}\label{subsec:var_rel_theor}
\noindent     In the following, we suggest a strategy how the \textit{relevance} of 
the single predictors can be compared for the different approaches. 
For this purpose, we first introduce for each approach the standard measure
that is typically used in the literature  and then explain, how these measures can be normalized
and how categorical predictors can be treated, to achieve a good degree of comparability.

\subsubsection*{Random Forest}
\noindent    The considered variable relevance $VarRel^{RF}$ for a random forest model is based on 
the method described by \citet{breiman_random}, namely the Variable Importance (VI)\footnote{This is often also called Gini-Importance}.
For a feature $X_i$, the VI considered here is defined as the sum of the weighted impurity decreases 
$p(t)\Delta i(s_t,t)$ averaged over all nodes in which $X_i$ is used as a partitioning feature:
\begin{align*}
VI^{RF}_i = \frac{1}{\text{ntree}} \sum_T \sum_{t \in T:v(s_t)=X_i} p(t) \Delta i(s_t,t),
\end{align*}
with $p(t) = \frac{n_t}{n}$ denoting the fraction of all observations assigned to node $t$, $\Delta i(s_t,t)$ 
the impurity decrease of an interior node $t$, and $v(s_t)$ the feature used to split node $t$.
Precise definitions of impurity decrease as well as the concept of impurity 
can be found in \citet{breiman_random} and \citet{elements_stat_learning} respectively.
The variable relevance of the $i$-th variable is then defined as
\begin{align}\label{eq:scale}
VarRel^{RF}_i := \frac{VI^{RF}_i}{\sum^p_{k=1} VI^{RF}_k}\,,
\end{align}
i.e.\ the relative proportion of the total VI of a model.

\subsubsection*{XGBoost}
\noindent   The variable relevance $VarRel^{XGB}$ for an XGBoost model is directly related 
to the Gini importance of a single classification tree. Let $VI_m(X_i)$ be the Gini importance of 
the $k$th feature of the $m$th tree of an XGBoost model. Then, the Gini importance of an 
XGBoost model is the average of all Gini importances of the trees computed in boosting, i.e.\
\begin{align*}
VI^{XGB}_i = \frac{1}{M} \sum_{m=1}^M VI_m(X_i)\,.
\end{align*}
In the case of a numerical variable, the variable relevance corresponds directly to the Gini importance.
For a categorical variable, however, we need to proceed somewhat differently. We assume that it has 
been passed to the model by dummy encoding. If $X_i$ is a categorical variable with $l$ different levels, 
the dummy variables $X_{i_{1}},\ldots,X_{i_{l}}$ are passed to the XGBoost procedure.
Then, we define a preliminary variable relevance via
\begin{align}
\widetilde{VarRel}^{XGB}_i = \begin{cases}
VI^{XGB}_i, & \text{if $X_i$ is numeric}, \\
\max\limits_{r \in \{i_1,\ldots,i_l\}}{VI^{XGB}_{r}}, & \text{if $X_i$ is a factor}.
\end{cases}
\label{eq:preVarRel}
\end{align}
We believe that the maximal Gini importance of the dummies corresponding to a categorical variable
serves as a good proxy for the factor's total Gini importance and makes it well comparable to the Gini importance
of metric predictors. Another possible option would be for example to consider the average over 
all dummy Gini importances of the factor.
Scaling \eqref{eq:preVarRel} analogously to \eqref{eq:scale} then yields the 
variable relevance for the $i$-th variable
\begin{align}\label{eq:scale2}
VarRel^{XGB}_i = \frac{\widetilde{VarRel}^{XGB}_i}{\sum^p_{k=1} \widetilde{VarRel}^{XGB}_k}\,.
\end{align}

\subsubsection*{Elastic Net}
\noindent The variable relevance for an elastic net model is based on 
the coefficient estimates used. Let $\hat\beta_1,\ldots,\hat\beta_p$ denote the estimated
coefficients of such a model (corresponding to standardized covariates). 
Analogous to the case of the XGBoost model, for the calculation of the variable relevance, 
we have to distinguish between numerical and categorical variables.
In the case of a numerical variable, we define the variable relevance simply as the absolute value of 
the calculated $\beta$ coefficient. In the case of a categorical variable, we again assume that it is passed 
to the model by dummy coding. Similar to before, the variable relevance can then be defined as
\begin{align}
\widetilde{VarRel}^{ElaNet}_i = \begin{cases}
|\beta_i|, & \text{if $X_i$ is numeric}, \\
\sqrt{l}*\sqrt{\sum\limits_{t=1}^l  \beta_{i_{t}}^2 }, & \text{if $X_i$ is a factor}.
\end{cases}\label{eq:VarRelElastic}
\end{align}
This way, the calculation of the variable relevance of a categorical variable 
was implemented following the idea of the Group LASSO procedure from \citet{meier2008group}.
Scaling \eqref{eq:VarRelElastic} analogously to \eqref{eq:scale} and \eqref{eq:scale2} 
then results in the variable relevance for the $i$-th variable.
\begin{align}
VarRel^{ElaNet}_i = \frac{\widetilde{VarRel}^{ElaNet}_i}{\sum^p_{k=1} \widetilde{VarRel}^{ElaNet}_k}.
\end{align}

\section{Results}\label{sec:results}
\noindent The analysis of the data described in Chapter~\ref{sec:data} was carried out in several steps. 
We start with some exploratory analyses in which we perliminarily examined relationships between individual variables.
Furthermore, we built models and tuned them with respect to their hyper-parameters. 
    
The main focus in our statistical analyses is the influence of the variables on individual contract cancellation behavior, 
which yields important insights into the factors affecting the lapse of life insurance contracts. 
Finally, these findings are presented graphically and compared with each other.

\subsection{Explorative analyses}\label{epl_an}
\noindent From a large number of results obtained in the exploratory analyses on the data sets of 
the ELI and the private pension insurance, as expected the descriptive results concerning the influence of the remaining 
contract duration and the sum insured turned out to be most informative. 

\subsubsection*{Remaining contract duration}
\begin{figure}
\centering
\includegraphics[width=1.05\linewidth]{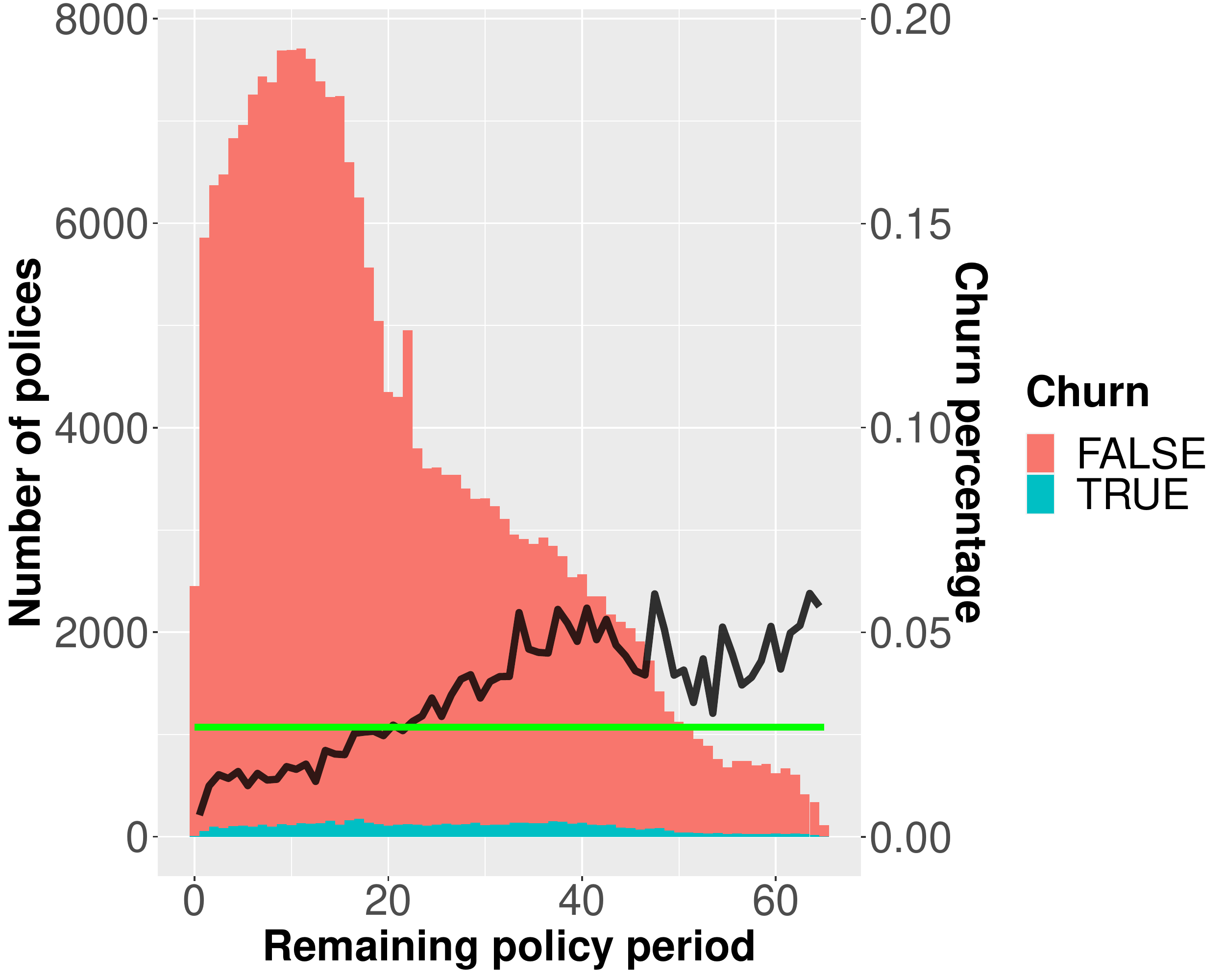}
\caption{Number of canceled and non-cancelled contracts depending on the remaining contract term; 
 the black line represents the lapse rate depending on the remaining contract duration, the green line represents 
 the lapse rate of the entire annuity data set.} \label{fig:laufzeit_rest_anteile}
\end{figure}

\noindent Figure~\ref{fig:laufzeit_rest_anteile} shows the absolute numbers of canceled and non-cancelled contracts 
for private pensions and the corresponding cancellation rates depending on the remaining duration of the contract.
It can be seen that the longer the remaining duration of the contract, the higher the cancellation rate.
Conversely, the closer the expiration date of an insurance policy, the less likely it is to be canceled.
These findings are not unusual, as life insurance policies are often canceled rather early, i.e.\ in the first few years of the contract. 
Various reasons play a role here, such as a wrong decision in case it turns out that the insurance 
is not needed after all, or the availability of a better offer, which often leads to just minor financial losses 
when the switch is made in the early years.
The longer a customer is holding a contract, the less likely she or he terminates it, 
since the benefit resulting from the contract is getting closer and closer. 
Finally, an economically acting individual would not cancel the contract after a longer contract period, 
because the financial losses become disproportionately high. 
Above all, a financial grievance could explain such late cancellations.
		    
\subsubsection*{Insurance Sum}
\begin{figure}[h!]
\centering
 \includegraphics[width=1.05\linewidth]{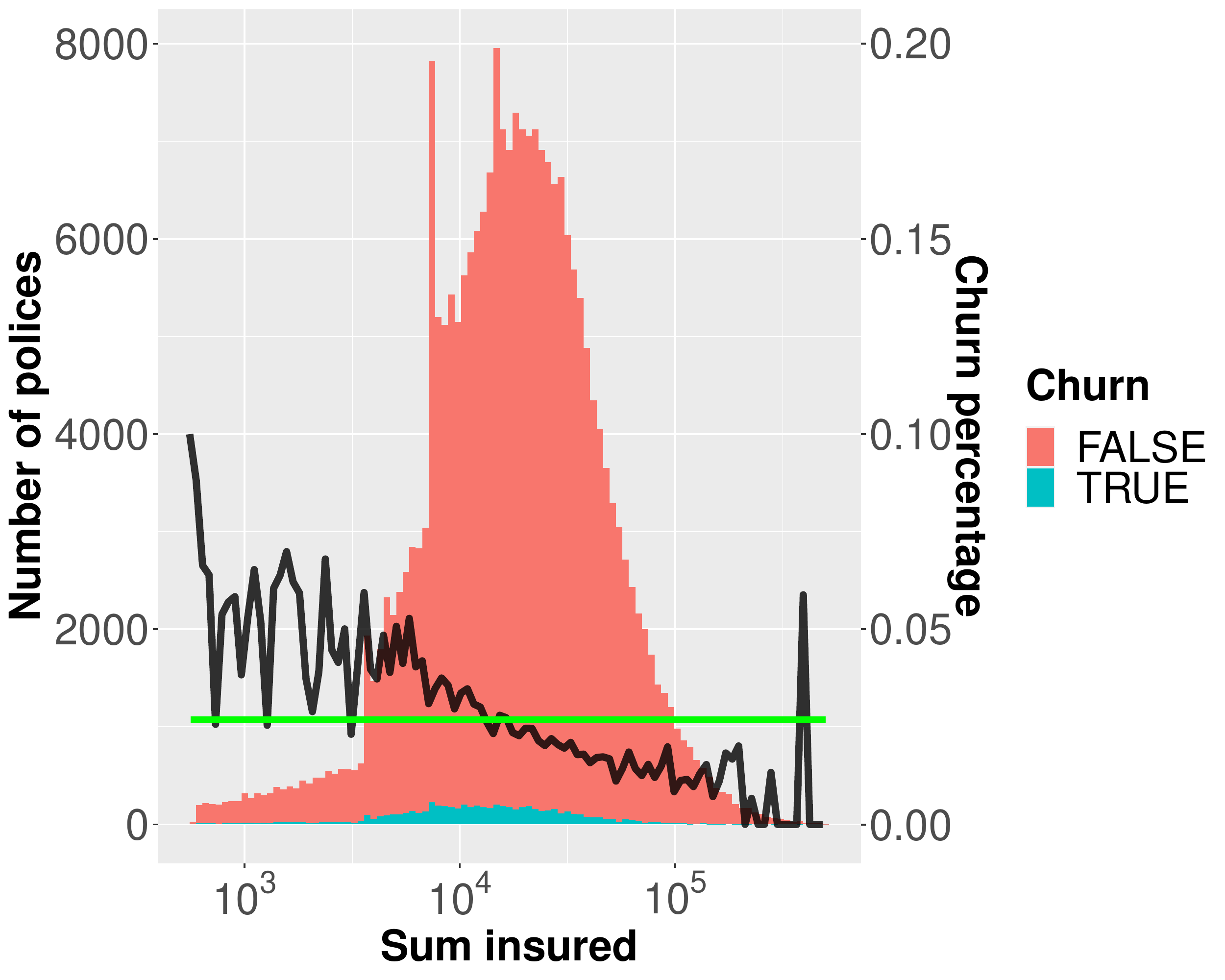}
\caption{Number of canceled and non-cancelled contracts depending on the sum insured. 
The black line represents the lapse rate as a function of the sum insured, the green line the lapse rate of the entire annuity data set.
The x-axis is plotted logarithmically.}
\label{fig:vsum}
\end{figure}

\noindent  As can be seen in Figure~\ref{fig:vsum}, the higher the sum insured, 
the lower the proportion of canceled contracts. From an economic point of view,
this makes sense, because a higher sum insured would 
lead to higher financial losses in the event of cancellation. 
In the case of private pension insurance, a low sum insured leads to a low monthly 
pension payment, which makes it less difficult to cancel the contract and invest the money elsewhere.

\subsection{Findings of Tuning}
\noindent  After the exploratory examination of the data, different settings of the 
models are investigated, which should finally provide the best model for each method.
        
Typically, for a random forest model, the more trees are trained, the better 
the model performance of such a model.
Table~\ref{tab:rf_ntree} confirms this statement, because the higher the value of the hyper-parameter 
\textit{ntree}, the better the model fit. On the other hand, the computational cost of a larger model compared 
to a smaller model is not necessarily proportional to the improvement in quality. 
Specifically, a random forest model using 300 trees will perform better, but not significantly better 
than the same model using only 50 trees. Nevertheless, the model must not be structured too simple.
As can be seen in Table~\ref{tab:rf_ntree}, a random forest model that uses only 10 trees performs 
substantially worse than a model with 50 trees.

\begin{table}[htp]
\begin{scriptsize}
\begin{tabulary}{13.7cm}[h]{|LLLL|LLLLLLLL|} \hline
 osw.rate & ntree & ntry & nodesize & auc.te & auc.tr & bac.te & bac.tr & br.te & br.tr & f1.te & f1.tr  \\\hline
 36 & 10 & 4 & 5000 & 0,7118 & 0,7454 & 0,6646 & 0,6854 & 0,2588 & 0,2394 & 0,0954 & 0,6927 \\ \hline
36 & 50 & 4 & 5000 & 0,7250 & 0,7594 & 0,6690 & 0,6882 & 0,2502 & 0,2302 & 0,0962 & 0,6972 \\\hline
36 & 300 & 4 & 5000 & 0,7273 & 0,7640 & 0,6693 & 0,6890 & 0,2480 & 0,2285 & 0,0965 & 0,6977 \\
\hline
\end{tabulary} 
\end{scriptsize}
\caption{Performance difference of a random forest model depending on the hyper-parameter \textit{ntree}. 
Exemplarily shown for the data set of the private pension; \textit{tr} stands for training data and \textit{te} for test data.}  
\label{tab:rf_ntree}
\end{table}

Another important aspect of this analysis is the handling or correction of the unbalanced target variable, 
i.e.\ the individual contract cancellation, in the data sets.         
Table~\ref{tab:rf_owsrate} shows exemplarily for the random forest model 
that the oversampling rate \textit{osw.rate}\footnote{This term 
stands for OversampleWrapper, a concept from \texttt{mlr}.}, i.e.\ by which factor the minority 
class is enlarged, has a direct positive impact on the model's goodness-of-fit.
Comparing the model without any adjustment of the training data, i.e.\ \textit{osw.rate} $=1$, 
to models in which an adjustment was made, the model fit is significantly worse.
It can be seen that an oversampling rate corresponding to the imbalance rate of a data set, i.e.\
\begin{align}
IR = \dfrac{N}{P}, \label{eq:IR}
\end{align}
yields the best performance. The imbalance rate describes the ratio of the minority to the majority class.
At the same time, we find that a too strong correction of the imbalance reduces the model quality again.

\begin{table}[htp] 
\begin{scriptsize}
\begin{tabulary}{13.7cm}[h]{|LLLL|LLLLLLLL|} \hline
 osw.rate & ntree & ntry & nodesize & auc.te & auc.tr & bac.te & bac.tr & br.te & br.tr & f1.te & f1.tr  \\\hline
 1 & 50 & 4 & 2500 & 0,5064 & 0,5244 & 0,5000 & 0,5000 & 0,0268 & 0,0267 & 0,0000 & 0,0000 \\ \hline
18 & 50 & 4 & 2500 & 0,7150 & 0,7650 & 0,6071 & 0,6323 & 0,0726 & 0,2072 & 0,1386 & 0,4579 \\\hline
36 & 50 & 4 & 2500 & 0,7282 & 0,7920 & 0,6693 & 0,7116 & 0,2301 & 0,2063 & 0,0985 & 0,7209 \\
\hline
54 & 50 & 4 & 2500 & 0,7203 & 0,8188 & 0,6599 & 0,7002 & 0,3468 & 0,1898 & 0,0828 & 0,8011 \\
\hline
\end{tabulary}	  	 
\end{scriptsize}	     
\caption{Performance difference of a random forest model depending on the hyper-parameter \textit{osw.rate}. 
Exemplarily shown for the data set of the private pension; \textit{tr} stands for training data and \textit{te} for test data.}  
\label{tab:rf_owsrate}
\end{table}

\begin{figure}[h!]
\centering
\includegraphics[width=0.7\linewidth]{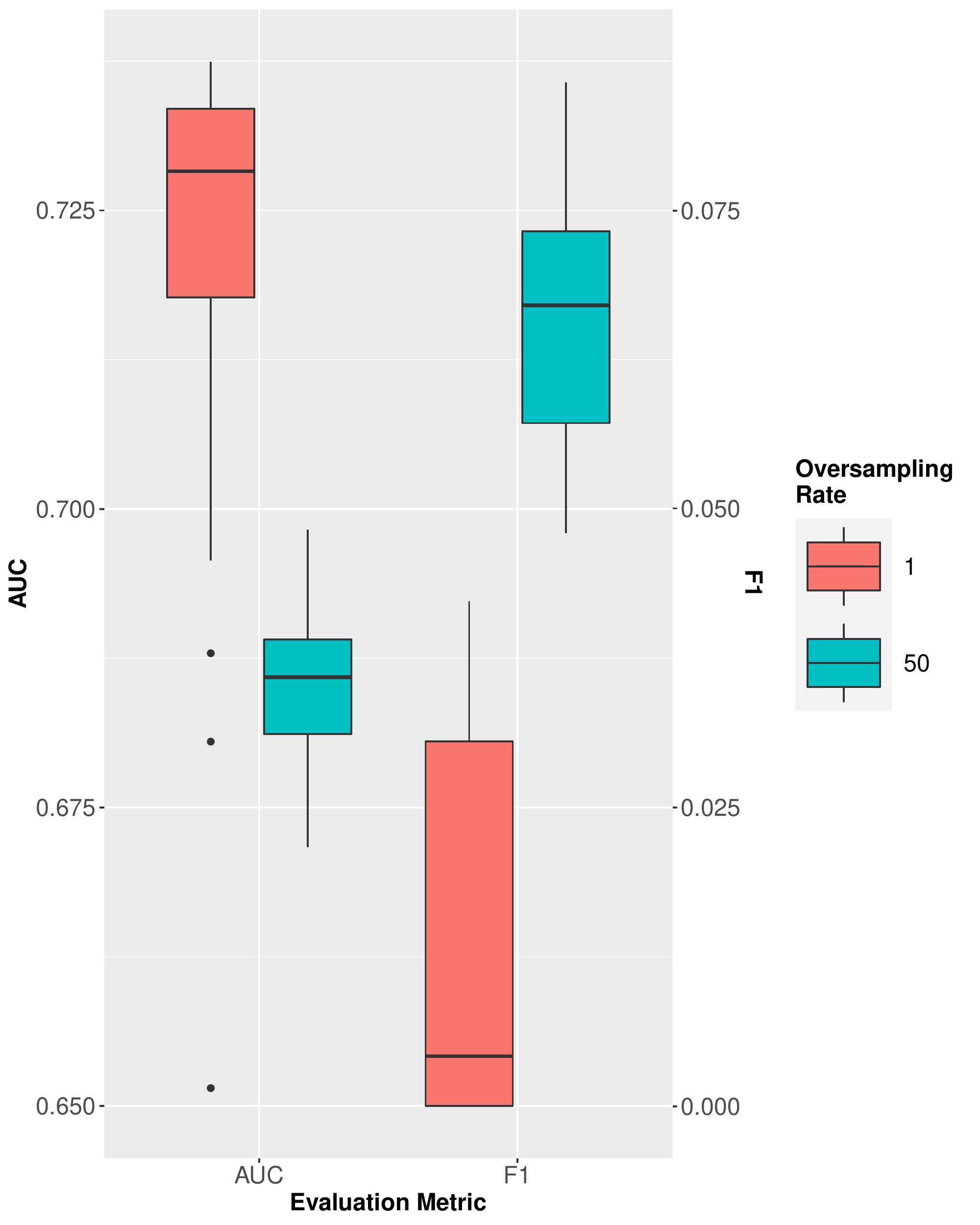}
\caption{Performance of XGBoost models generated in tuning depending on the hyper-parameter \textit{sw.rate}}
\label{fig:xgb_swrate_auc_f1}
\end{figure}

We additionally compared the AUC and F1 performance metrics, which were 
calculated on external test data during the tuning of an XGBoost model for the ELI data set, 
see Figure~\ref{fig:xgb_swrate_auc_f1}. It is noticeable that XGBoost models without an oversampling adjustment 
in training achieve better AUC values than XGBoost models with an oversampling adjustment 
close to the imbalance rate \eqref{eq:IR}. In addition, XGBoost models without an oversampling adjustment 
in training achieved a worse performance with respect to the F1 measure compared to models that use 
an oversampling rate corresponding to the imbalance rate \eqref{eq:IR}.
            
In summary, different oversampling rates lead to different model performances and, 
depending on the used learning procedure and the employed oversampling rate, 
different models are to be preferred.

\subsection{Performance}    
\noindent In this study, the employed models are based on different classification methods. 
The evaluation of the results of these models is done as described in Chapter~\ref{subsec:evaluation}.
During the parameter tuning of these models, different parameter settings were tried and for 
each of them the corresponding test and training performances with respect to both 
performance measures were calculated. 

Table~\ref{tab:best_performance} shows the best performance on external test data
for each classification method and both goodness-of-fit measures. First of all, it can be observed 
that none of the models achieves a remarkable predictive performance, 
regardless of which data set is analyzed. 
            
Comparing the values obtained in Table~\ref{tab:best_performance} with the benchmarks 
from \citet{gold2020fighting}, it can be concluded that the models achieve a moderate predictive performance. 
The goal would be to achieve e.g.\ an AUC of $0.8-0.85$, see \citet{gold2020fighting}.
             
A more detailed analysis reveals differences between the different learning methods. 
On the one hand, both the random forest and the XGBoost method achieve the best test performance 
values with respect to the measures AUC and, on the other hand, the Logit model shows the worst performance 
for all goodness-of-fit measures. For this reason, the logit models and CART models are no longer considered 
for the analysis of the variables in the further course.
In addition to the analysis of the best test performance values, the other models will now be examined in more detail.  

\begin{center}       
\begin{table}[htp]
\begin{tabulary}{15cm}[h]{|C|C|C|C|} \hline
& Approach & AUC & F1 \\ \hline\hline
Endowment life insurance & Logit & 0.6440 & 0.0522 \\ \hline
& Elastic Net & 0.6931 & 0.0985 \\ \hline
& CART & 0.7070 & 0.1002 \\ \hline
& Random Forest & 0.7221 & 0.1124 \\ \hline
& XGBoost & 0.7335 & 0.0779 \\ \hline\hline
Private pension insurance & Logit & 0.6666 & 0.0946 \\ \hline
& Elastic Net & 0.7115 & 0.1226 \\ \hline
& CART & 0,7106 & 0.1263 \\ \hline
& Random Forest & 0.7333 & 0.1529 \\ \hline
& XGBoost & 0.7367 & 0.1405 \\ \hline
\end{tabulary}
\caption{Overview of the best performances for different classification
methods on external test data and for two different real data sets.}  
\label{tab:best_performance}
\end{table}
\end{center}

\subsubsection{ROC-plots}  
\begin{figure}[htp]
\centering
\begin{subfigure}{.57\textwidth}
\centering
\includegraphics[width=\linewidth]{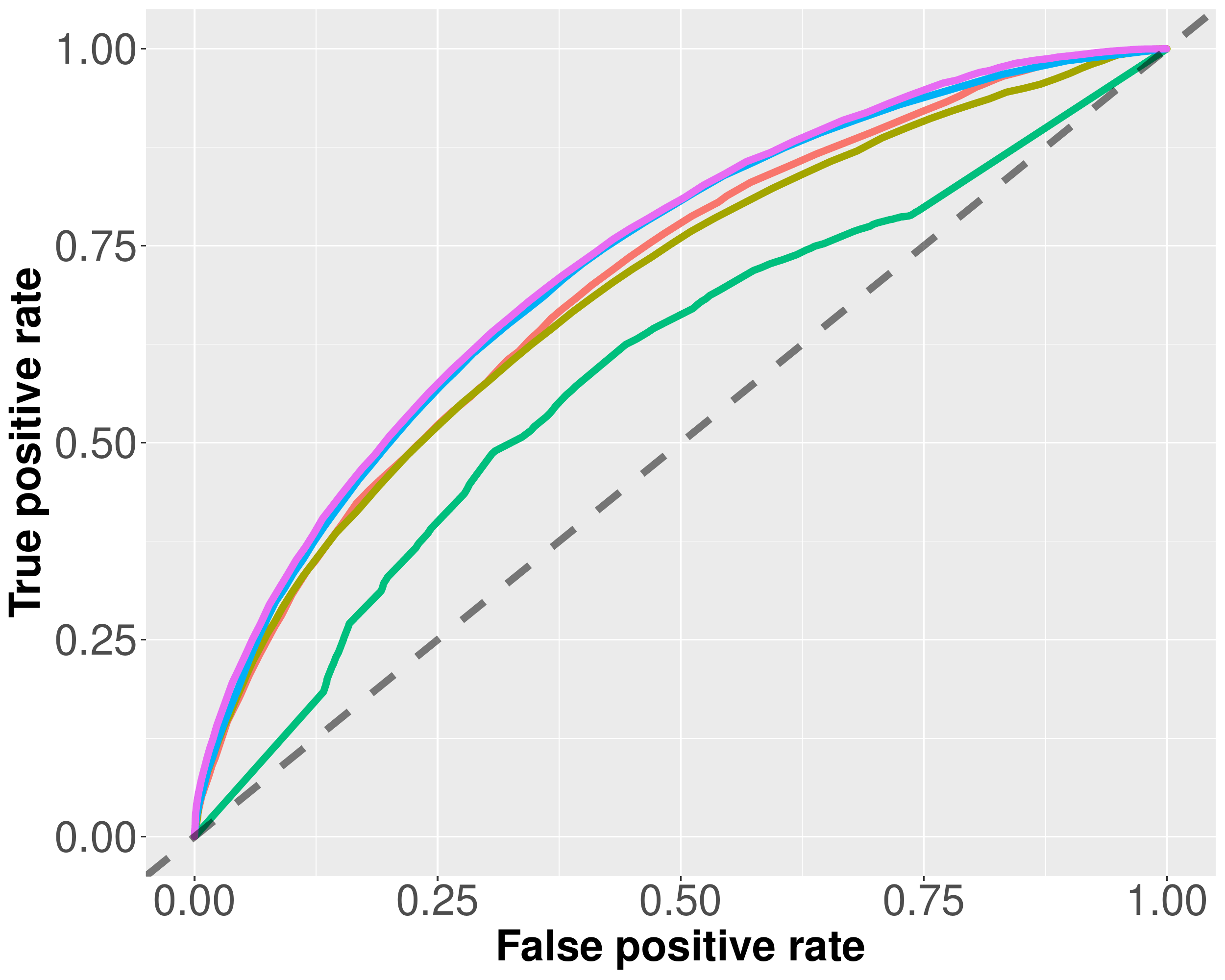}
\caption{ROC curves}
\label{OS_roc_klv}
\end{subfigure}
\vspace{1.5em}

\begin{subfigure}{.57\textwidth}
\centering
\includegraphics[width=\linewidth]{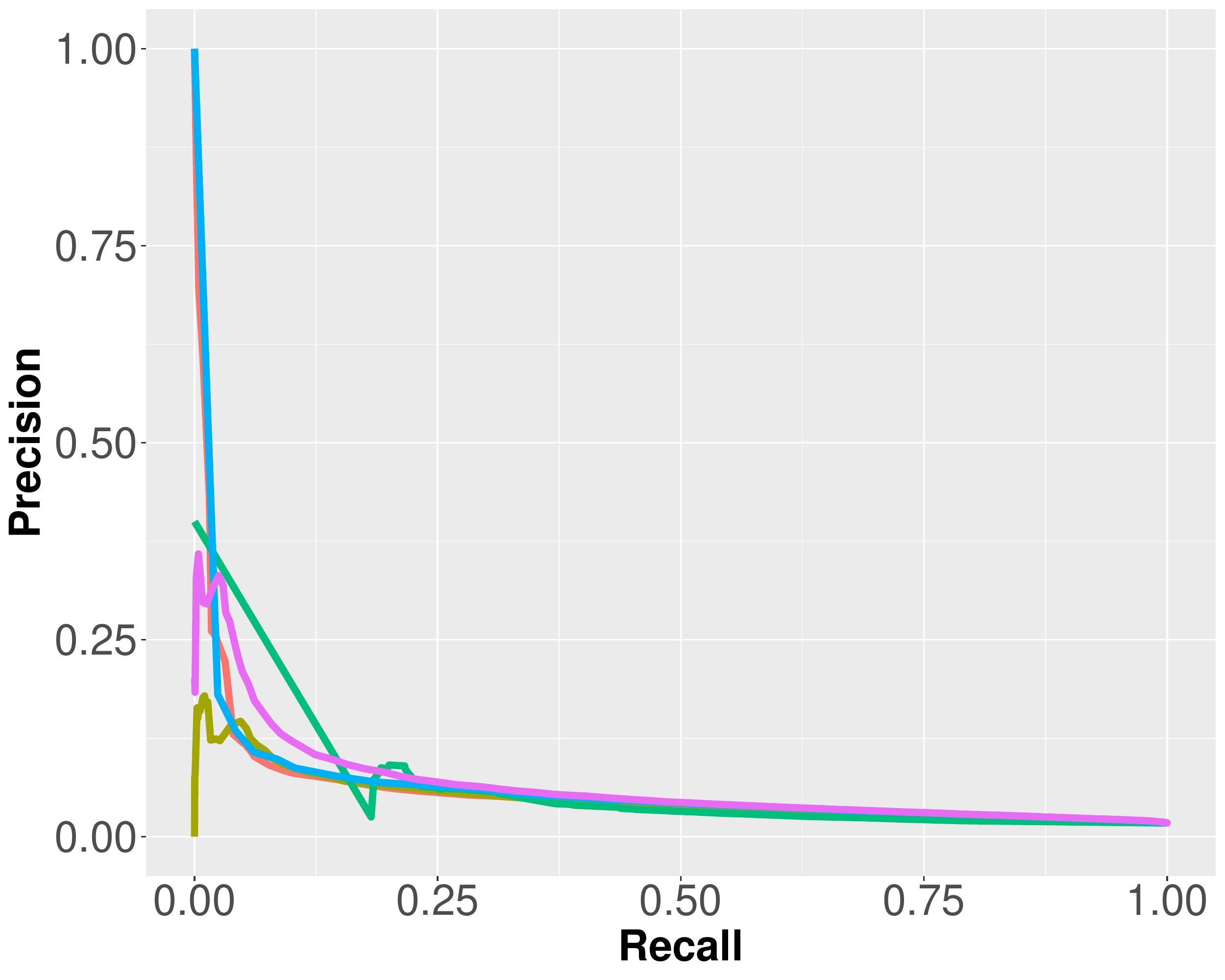}
\caption{PR curves}
\label{OS_pr_klv}
\end{subfigure}	 

\begin{subfigure}{0.6\textwidth}
\centering
\includegraphics[width=\linewidth]{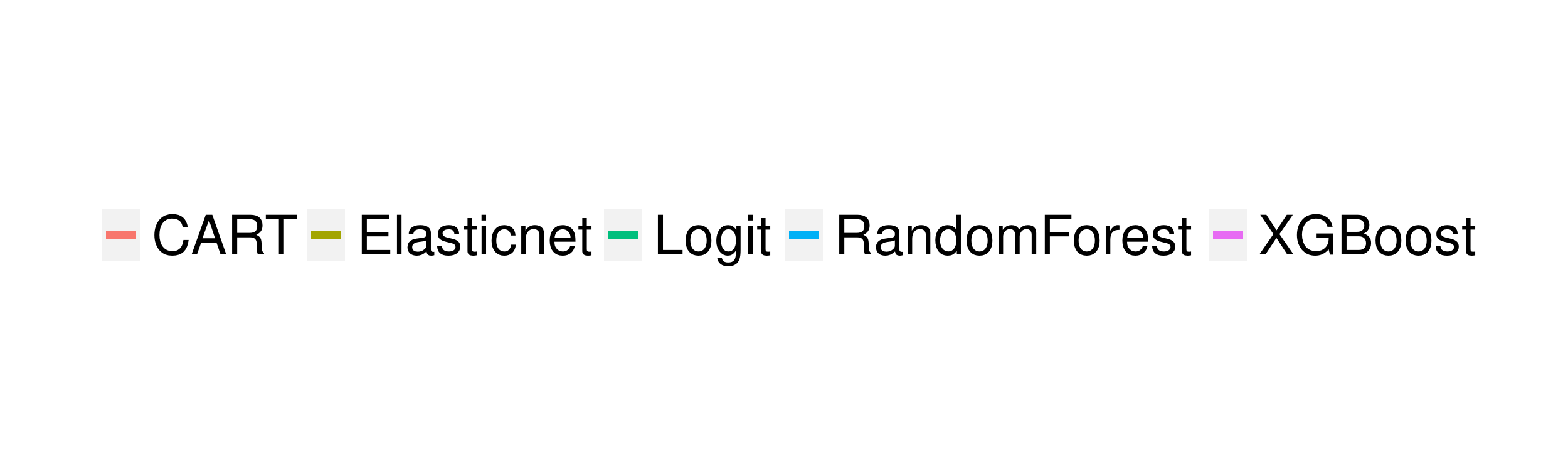}
\end{subfigure}
\caption{
ROC and Precision-Recall curves for the endowment life insurance data set for the models that are optimal in 
terms of AUC. The models were computed using random oversampling as well as 10-fold cross-validation. \\
In Figures~\ref{OS_roc_klv} and \ref{OS_pr_klv}, the corresponding ROC and Precision-Recall curves 
(regarding cross-validation) are shown in an aggregated form.}	  
\label{opt_auc_klv}
\end{figure}

\begin{figure}[htp]
\centering
\begin{subfigure}{.57\textwidth}
\centering
\includegraphics[width=\linewidth]{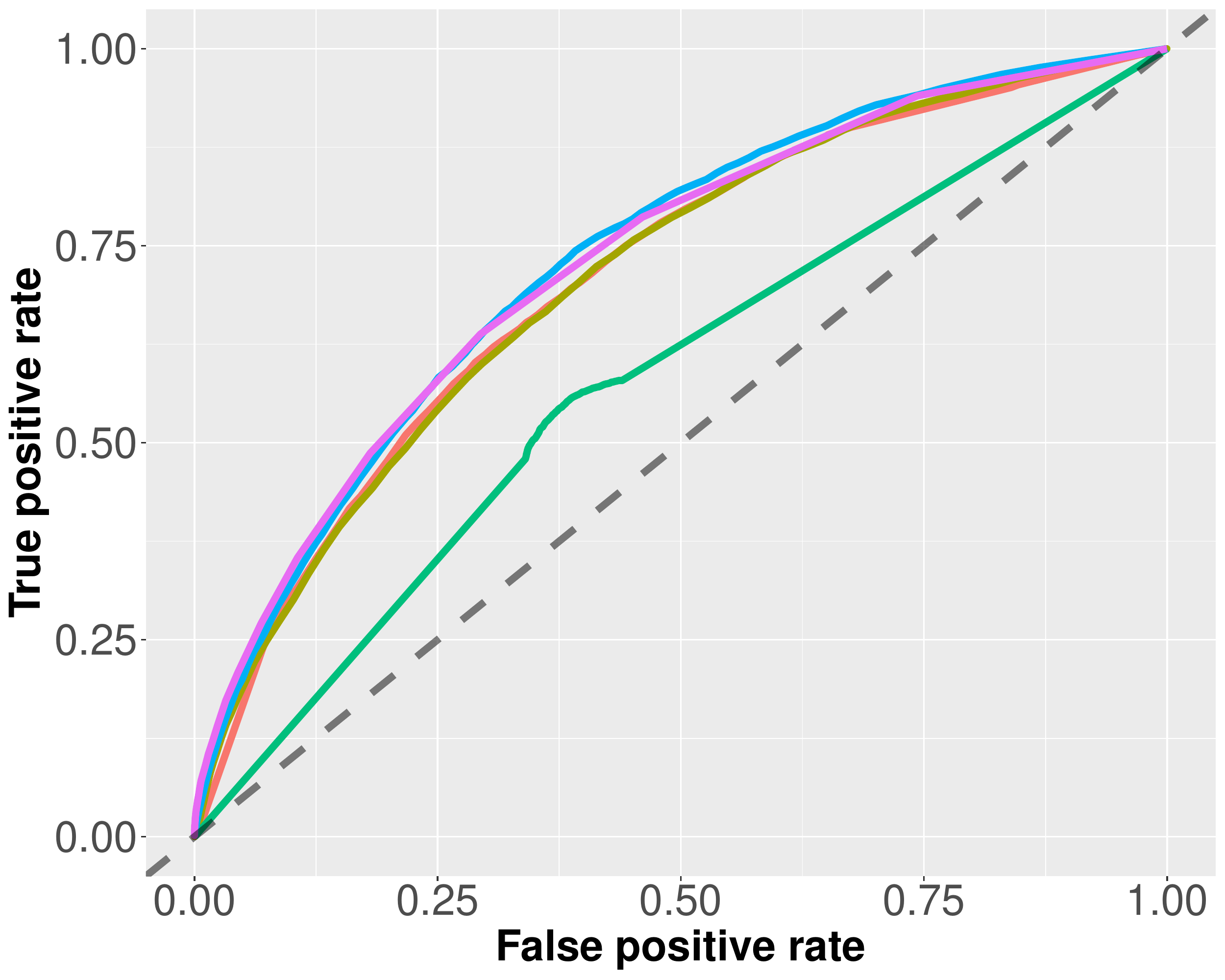}
\caption{ROC-kurven}
\label{OS_roc}
\end{subfigure}
\vspace{1.5em}

\begin{subfigure}{.57\textwidth}
\centering
\includegraphics[width=\linewidth]{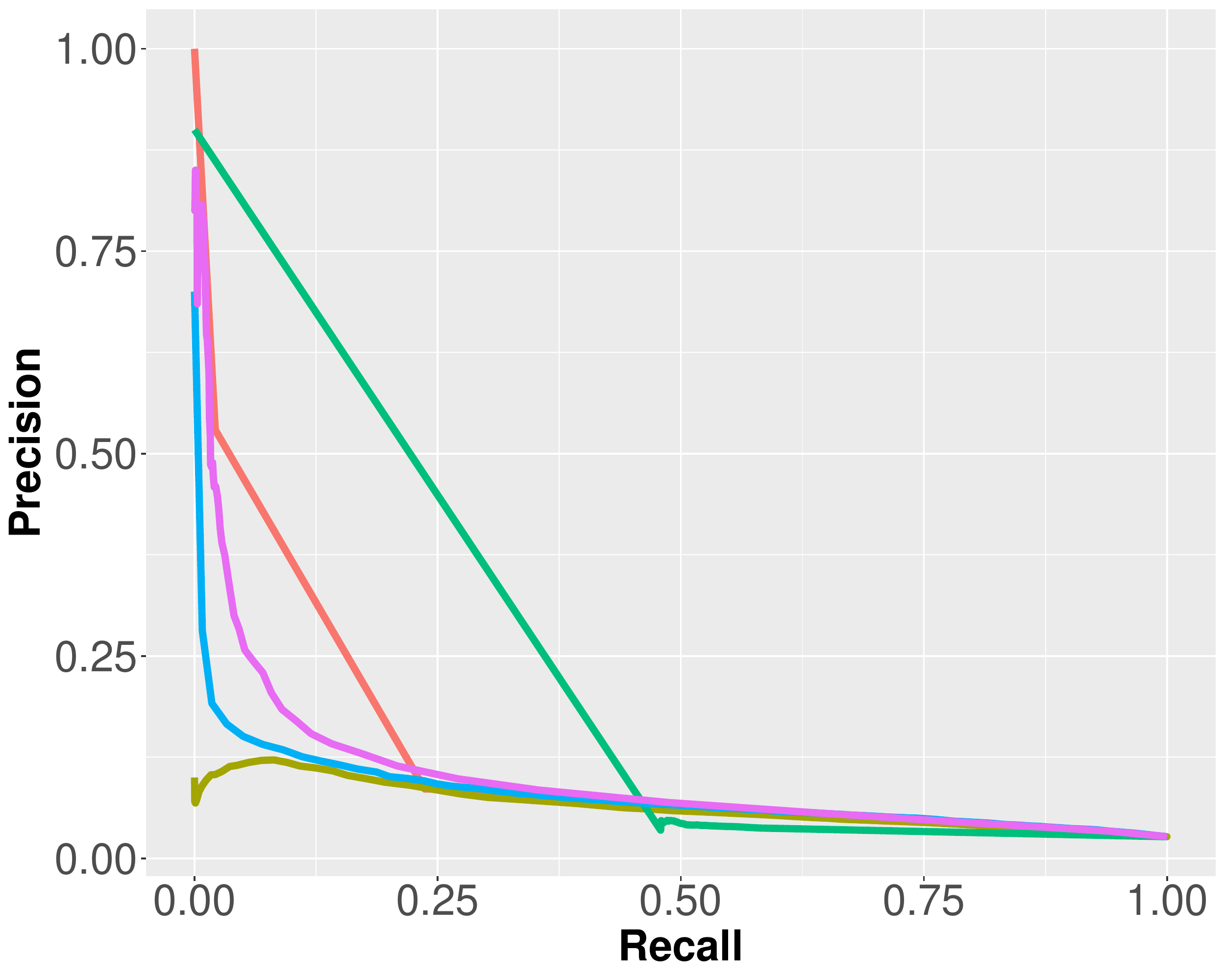}
\caption{PR-Kurven}
\label{OS_pr}
\end{subfigure} 
		    
\begin{subfigure}[t]{0.6\textwidth}
\centering
\includegraphics[width=\linewidth]{Legend.pdf}
\end{subfigure}
\caption{ROC and Precision-Recall curves for the private pension data set for the models 
that are optimal in terms of AUC. The models were computed using random oversampling as well as 10-fold cross-validation. \\
In Figures~\ref{OS_roc_klv} and \ref{OS_pr_klv} the corresponding ROC and Precision-Recall curves 
(regarding cross-validation) are shown  in an aggregated form.}	  
\label{opt_auc}
\end{figure}

\noindent Figures~\ref{opt_auc_klv} and \ref{opt_auc} show the ROC and Precison/Recall 
curves for the models that are optimal with respect to AUC. 
In Figures~\ref{OS_roc_klv}, \ref{OS_pr_klv}, \ref{OS_roc}, and \ref{OS_pr}, 
the models were fitted in combination with random oversampling.

The course of the ROC curves neither indicates a solely random, and thus very bad, 
nor a very good model performance. Furthermore, all shown ROC curves, except for those corresponding to 
the logit models, are rather close to each other and show a very similar behavior, whereas the curves of the random forest and 
XGBoost models show a slightly better performance. 

Looking at the Precision-Recall curves of the learning methods, all models show a rather 
poor performance for the (underrepresented) positive observations, since all curves run along the lower left corner. 


\subsubsection{Variable Relevance}\label{var_rel_erg}
\begin{figure*}
\centering
\includegraphics[scale=0.55,angle = 90]{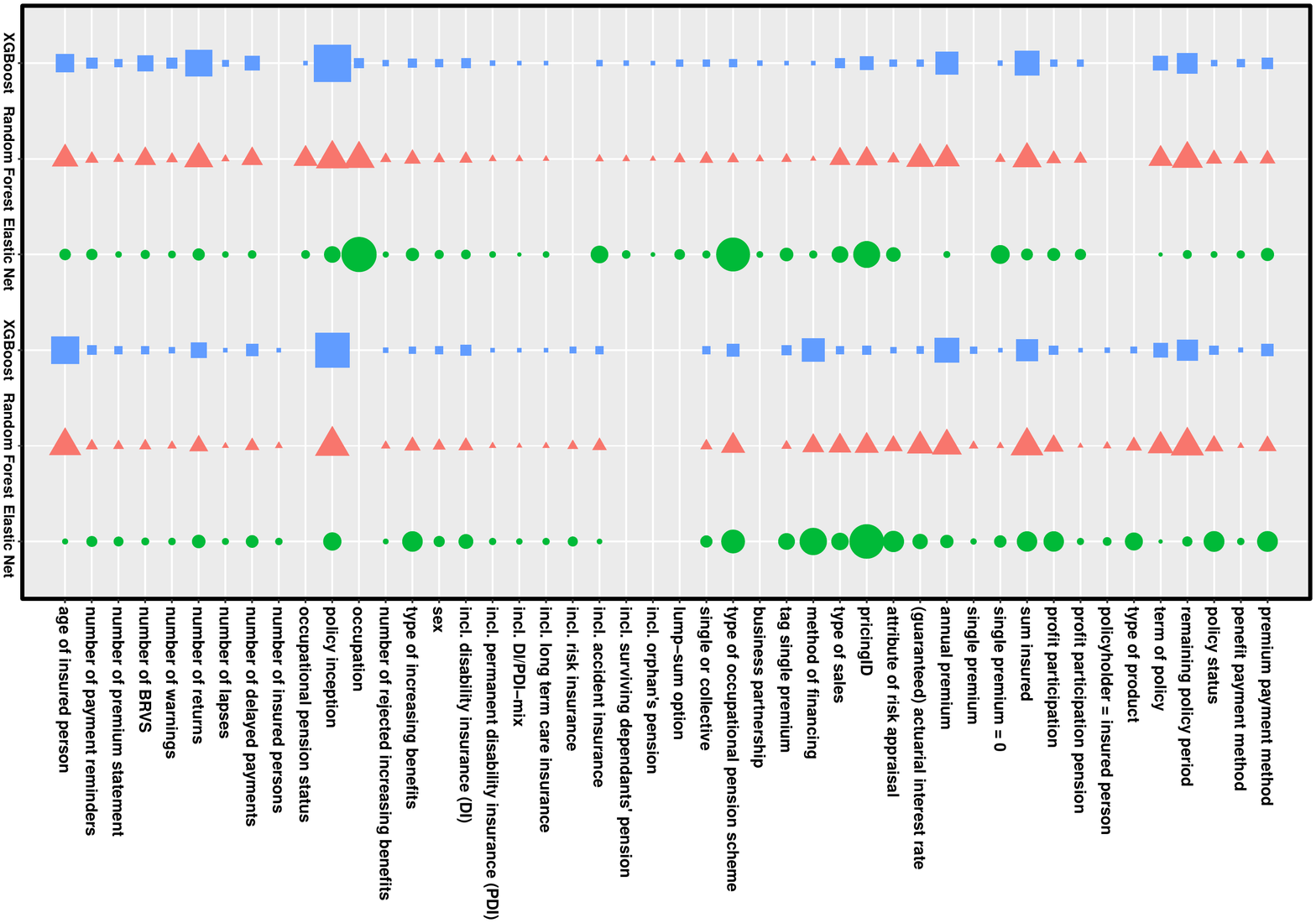}
\caption{Variable relevances for three different classification approaches: XGBoost (blue \& left), 
random forest (red \& middle) and  elastic net (green \& right). The first three columns 
correspond to the data set of the private pension, columns 4-6 to the endowment insurance 
data set.}
\label{var_rel}
\end{figure*}

 \noindent In addition to the ROC and Precision-Recall curves, methods for calculating the 
 {\it Variable Relevance} were introduced in Section~\ref{subsec:var_rel_theor}. 
 With the help of these characteristics, the major research questions of this project can be answered, 
 which concerns the determination of the concrete reasons and causes of lapse behavior for life insurance contracts.
To answer this question, we examine in more detail those elastic net, random forest, and 
XGBoost models that are optimal with respect to AUC. For this purpose, we consider the corresponding variable relevance values 
of both data sets, which are displayed in Figure~\ref{var_rel}. The graph can be read as follows:
The larger the respective symbol, the larger the corresponding value of variable relevance and 
thus the influence on the predictive power of a model.

The predictors identified as most important via the concept of the introduced variable relevance 
methods can generally be grouped into three types for both examined data sets: Time-related influences, 
contract content, and information from the collection system.
            
\begin{itemize}
\item  The temporal influences include information on the \texttt{policy inception}, the \texttt{remaining policy period} of a contract 
part, and the  \texttt{age of insured person}.
\item Important features related to the contract content are, for example, the contractually agreed 
\texttt{sum insured} and the \texttt{annual premium} paid. Especially for the private pension data set, 
also the \texttt{occupation} has been identified as an important feature.
\item With regard to the information used from the collection system, the \texttt{number of returns} was found to be relevant, 
especially in the context of the private pension insurance data. Here, a return describes a payment request by the insurance 
company that has not been answered by the customer.
\end{itemize}
Principally, it turns out that all three approaches show a rather high agreement on both 
regarded data sets in terms of which variables are more or less relevant. We will now discuss 
our findings regarding the variable relevance of the different features in more detail.
Using the variable relevance of all model types listed in Figure~\ref{var_rel}, it can be seen, 
for example, that the presence of a particular supplementary insurance will have little impact on 
the predictive validity, see characteristics of type "\texttt{incl.~xxx}".

Some other variables such as the \texttt{PricingID}, the \texttt{Method of financing}, the BAV indicator 
(\texttt{type of occupational pension scheme}), or the \texttt{profit participation} turn out to be 
quite important in some models, whereas they appear to be of little relevance in others. 
It must be emphasized, however, that most of the predictors that are relevant for the predictive performance
within one model also exhibit some importance in other models.
For a large subset of these features, this statement can even be made for both data sets, which 
allows us to infer that these features have a high influence on individual policy lapses for at least the 
regarded two product types of life insurance. It should also be emphasized that these findings confirm the aspects 
identified in Section~\ref{epl_an}, e.g.\ a strong influence of the remaining contract duration or the sum insured. 
            
All in all, however, the computed models unfortunately only achieve a limited degree of predictive accuracy.
Against this background, it is important to note that the predictors identified as particularly 
important by the variable relevance methods are not necessarily the most important predictors that generally exist.
In other words, other possibly important features have not been included in the data and models used in this study
(see Section~\ref{sec:conclusion} for a discussion of candidate covariates). 

\section{Conclusion and outlook}\label{sec:conclusion}
    %
\noindent	In the present study, we investigated how well the portfolio data of a large German 
life insurance company can be used to predict and explain individual contract cancellation behavior. 

For this purpose, data from endowment life insurance and private pensions were used and processed. 
Furthermore, some forecasting methods were presented in the manuscript, 
which were subsequently applied to these data sets. 
	
As a starting point the date 1/1/2018 was chosen. Then, the inventory data of the two products described above were used, 
which were combined with the cancellation data of the year 2018. Thus, it was possible to 
determine for each contract in the insurer's portfolio as of 1/1/2018 whether it had been 
terminated by lapse at the end of the year.  
	 
For each product, only the main insurance of the contract was considered. Information about the customer, time components, 
information about additional insurances and some other details from the collection system were added to the usual 
contract data. This resulted in data sets for both products with a sufficient number of features to apply different prediction methods. 
	 
In addition to classical logistic models, which included the standard logit model and an elastic net-regularized version of it, 
tree-based methods such as CART, random forest and XGBoost were also used. Based on these methods, 
a large number of models per insurance product was considered and the best model was selected based on tuning strategies. 
For the tree-based methods, parameters such as the number of trees, the depth of the trees, the size of the nodes and the 
oversampling rate were considered in the tuning process. The last tuning parameter was particularly relevant here, as 
life insurance contracts naturally are rarely cancelled and thus, there were few cancelled as opposed to non-cancelled contracts.  
This unbalancedness of our data sets was taken into account by using an oversampling approach, so-called 
random oversampling. 
	 
The analysis of the models using ROC and precision-recall plots clearly showed that the cancellation 
behavior of individual contracts could only be explained moderately well on the basis of the available data. 
None of the classification methods investigated in this project substantially outperformed the others or achieved
 a very satisfactory predictive accuracy. 
	 
The concept of variable relevance, which was used to analyze and interpret the individual contract cancellation 
behavior, finally showed that some temporal components, such as the start of the contract, the remaining duration of the contract or 
the age of the first insured person, but also certain contract contents, such as the sum insured, the annual premiums paid, the surplus system used, 
as well as information from the collection system, primarily the number of repayments, had a substantial influence on the prediction. 
Since the values of the model's goodness-of-fit are neither particularly bad or good, these findings should be viewed with some caution.

As an outlook and extension to this work, two major steps can be considered in further research: 
Incorporating additional data sources or using other models, such as e.g.\ boosting approaches or 
deep learning methods such as neural networks. 

\subsection*{Other data sources}
\noindent The incorporation of additional data sources always requires the inclusion of data protection regulations, 
so that both the extraction and further use of additional covariate data can be done in compliance with the law. 
Possible data sources that could provide essential information within the focus of this study are 
\begin{itemize}
\item[(a)] \textbf{The company's stored correspondence with the customer} \\~\\
 In the future, for example, information on customer satisfaction, on the cause of various steps taken, 
 or additional personal customer information could be extracted from various data sources available 
 to the life insurance company, e.g. via using text mining methods. This information could then be 
 added to the data records, thereby increasing the number of features, and thus the information content.
\item[(b)] \textbf{All existing customer contracts} \\~\\
 This additional information could provide insight into whether customers who hold 
 several different insurance contracts with the company have a different individual 
 contract cancellation behavior than customers who are not that deeply rooted with the company. 
 Changes such as cancellations or re-signings of other contracts may have a direct 
 impact on the contracts under investigation. 
\item[(c)] \textbf{Extraction of customer information from social media platforms} \\~\\
Data from social media platforms is a modern and widely used field of analysis in other 
industries to analyze changes in private relationships.
\end{itemize}

\subsection*{Other modeling approaches}
\noindent In this manuscript, solely classification methods have been considered. 
A possible extension could be to focus more on temporal effects in the future. 
Then, models from the field of time series analysis or time-to-event analysis could be considered. 
The fact that temporal information plays an important role in the models considered in this work 
could be interpreted as a first hint to indeed further pursue this idea.
One problem that arises, for example, for models from time series analysis is the creation 
of a meaningful data basis. The content of this data basis should be time series that show 
the historical course of a contract. However, since these are not always available in the desired 
quality, this turn out to be an insurmountable hurdle. 

In conclusion, it can be stated that statistical analyses in the field of insurance company portfolio 
management can be promising, but as also shown in the present manuscript, could arise some problems. 
Some of these problems can be addressed and potentially solved, while for others the effort involved 
in their solution is not commensurate with the associated added value. 

\bibliography{stornopaper.bib}

\end{document}